\documentclass[10pt,twocolumn,letterpaper]{article}

\usepackage[pagenumbers]{wacv} 

\usepackage{graphicx}
\usepackage{amsmath}
\usepackage{amssymb}
\usepackage{booktabs}

\usepackage{multirow}
\usepackage{soul}
\usepackage{pifont}
\usepackage{marvosym}

\usepackage[dvipsnames]{xcolor}
\usepackage{colortbl}
\definecolor{mygray}{gray}{.9}

\usepackage[pagebackref,breaklinks,colorlinks]{hyperref}

\usepackage[capitalize]{cleveref}
\crefname{section}{Sec.}{Secs.}
\Crefname{section}{Section}{Sections}
\Crefname{table}{Table}{Tables}
\crefname{table}{Tab.}{Tabs.}

\makeatletter
\renewcommand\paragraph{\@startsection{paragraph}{4}{\z@}%
  {1.8ex \@plus0.5ex \@minus.2ex}%
  {-1em}%
  {\normalfont\normalsize\bfseries}}
\makeatother

\newcommand{\cmark}{\ding{51}}
\newcommand{\xmark}{\ding{55}}

\newcommand{\T}{{\top}}

\newcommand{\kk}{\mathbf{k}}
\newcommand{\qq}{\mathbf{q}}
\newcommand{\ttt}{\mathbf{t}}
\newcommand{\vv}{\mathbf{v}}
\newcommand{\xx}{\mathbf{x}}
\newcommand{\XX}{\mathbf{X}}

\newcommand{\ZZ}{\mathbf{Z}}
\newcommand{\mmu}{\boldsymbol{\mu}}
\newcommand{\mbf}[1]{\mathbf{#1}}

\newcommand{\channel}{C}
\newcommand{\oursa}{\widetilde{\operatorname{SA}}}
\newcommand{\ourmethod}{NACLIP\xspace}
\newcommand{\og}{\textit{Neighbourhood Only}\xspace}
\newcommand{\ogshort}{N-Only\xspace}
\newcommand{\kksim}{\textit{Key-Key Similarity}\xspace}
\newcommand{\kksimshort}{KK-Sim\xspace}
\def\real{{\mathbb{R}}}
\newcommand{\softmax}{\mathrm{softmax}}

\begin{document}

\title{Pay Attention to Your Neighbours: \\Training-Free Open-Vocabulary Semantic Segmentation}

\author{\phantom{Jose Dolz} Sina Hajimiri\textsuperscript{\Letter} \qquad Ismail Ben Ayed \qquad Jose Dolz \phantom{Sina Hajimiri\textsuperscript{\Letter}} \\
ÉTS Montreal \\
\Letter \ {\tt \small seyed-mohammadsina.hajimiri.1@etsmtl.net}
}
\maketitle

\begin{abstract}
Despite the significant progress in deep learning for dense visual recognition problems, such as semantic segmentation, traditional methods are constrained by fixed class sets. Meanwhile, vision-language foundation models, such as CLIP, have showcased remarkable effectiveness in numerous zero-shot image-level tasks, owing to their robust generalizability.
Recently, a body of work has investigated utilizing these models in open-vocabulary semantic segmentation (OVSS). However, existing approaches often rely on impractical supervised pre-training or access to additional pre-trained networks. 
In this work, we propose a strong baseline for training-free OVSS, termed Neighbour-Aware CLIP (\ourmethod), representing a straightforward adaptation of CLIP tailored for this scenario. Our method enforces localization of patches in the self-attention of CLIP's vision transformer which, despite being crucial for dense prediction tasks, has been overlooked in the OVSS literature. 
By incorporating design choices favouring segmentation, our approach significantly improves performance without requiring additional data, auxiliary pre-trained networks, or extensive hyperparameter tuning, making it highly practical for real-world applications.
Experiments are performed on 8 popular semantic segmentation benchmarks, yielding state-of-the-art performance on most scenarios.
Our code is publicly available at \url{https://github.com/sinahmr/NACLIP}.
\end{abstract}

\section{Introduction} \label{sec:intro}

In recent years, we have witnessed remarkable progress of deep learning models in dense visual recognition tasks, such as semantic segmentation \cite{minaee2021image}. Nevertheless, a main limitation of these methods stems from the fixed set of classes available in the traditional training scenario. This hinders the deployment of these models in a myriad of real-world problems, where the number of visual concepts is far from being finite, and likely includes novel categories unseen during training. A straightforward solution would consist in collecting a large set of labeled images of each novel class to adapt the model. However, this solution is impractical in many aspects, from the cumbersome labeling process of additional images to the unrealistic adaptation of the model to each new class.

Open-vocabulary semantic segmentation (OVSS) has emerged as an appealing alternative to traditional close-set approaches, as it can handle novel categories that may not have been seen during training. Fueled by the zero-shot performance of vision-language models in visual recognition tasks, most recent OVSS approaches are inspired by the Contrastive Language-Image Pre-training (CLIP) \cite{clip} paradigm. A popular family of approaches integrates a \textit{fully-supervised} training step \cite{barsellotti2023enhancing,ghiasi2022scaling,liang2023open,xu2023open,xu2023side,xu2022simple,qin2023freeseg}, where pixel-wise masks from a limited set of categories are leveraged to transfer language-visual alignment at image level to pixel granularity. The pixel-level annotation assumption can be relaxed by considering a \textit{weakly-supervised} adaptation dataset, where only image-text pairs are accessible \cite{chen2023exploring,cai2023mixreorg,zhang2024uncovering,wu2023clipself,tcl,groupvit}. These methods, however, still require a substantially large set of annotations, which typically present a large overlap with the open-set categories in the test datasets (please see Table 1 in \cite{xu2023side}). Furthermore, the final model's performance may be biased towards the training dataset selected for adaptation.

To meet the requirements of real-world applications, where access to large labeled datasets is rare and one cannot anticipate novel classes, we focus in this work on \textit{training-free}, a more challenging and realistic scenario without access to additional data for adaptation. Due to their practical relevance, there exists a growing literature on these approaches \cite{maskclip,reco,wysoczanska2024clip,ovdiff,sclip,bousselham2023grounding,barsellotti2024fossil,clipsurgery,luo2024emergent,freesegdiff}, which leverages CLIP \cite{clip} as the source of knowledge. However, some also utilize additional pre-trained networks like MoCov2 \cite{he2020momentum}, DeiT \cite{naseer2021intriguing}, or unsupervised object localization approaches \cite{simeoni2023unsupervised} trained on extra datasets \cite{reco,wysoczanska2024clip,ovdiff}, or pre-trained text-conditional generative models (\eg, stable diffusion) to generate multiple additional images for novel concepts \cite{barsellotti2024fossil,ovdiff,freesegdiff}. We argue that methods using these auxiliary pre-trained models, having been trained on significant extra data and introducing additional weights, cannot be fairly compared to \textit{training-free} methods that use only CLIP.
Furthermore, some prior works present practical limitations such as tuning many hyperparameters \cite{barsellotti2024fossil,luo2024emergent}, whose values change across datasets \cite{barsellotti2024fossil}, or integrating a weakly supervised reward function for hyperparameter tuning \cite{luo2024emergent}.
Thus, devising novel alternatives that relax these requirements and present simple solutions in a \emph{more realistic training-free} manner is paramount to deploying these methods in real-world problems.

\begin{figure*}[ht!]
    \centering
    \includegraphics[width=0.8\textwidth]{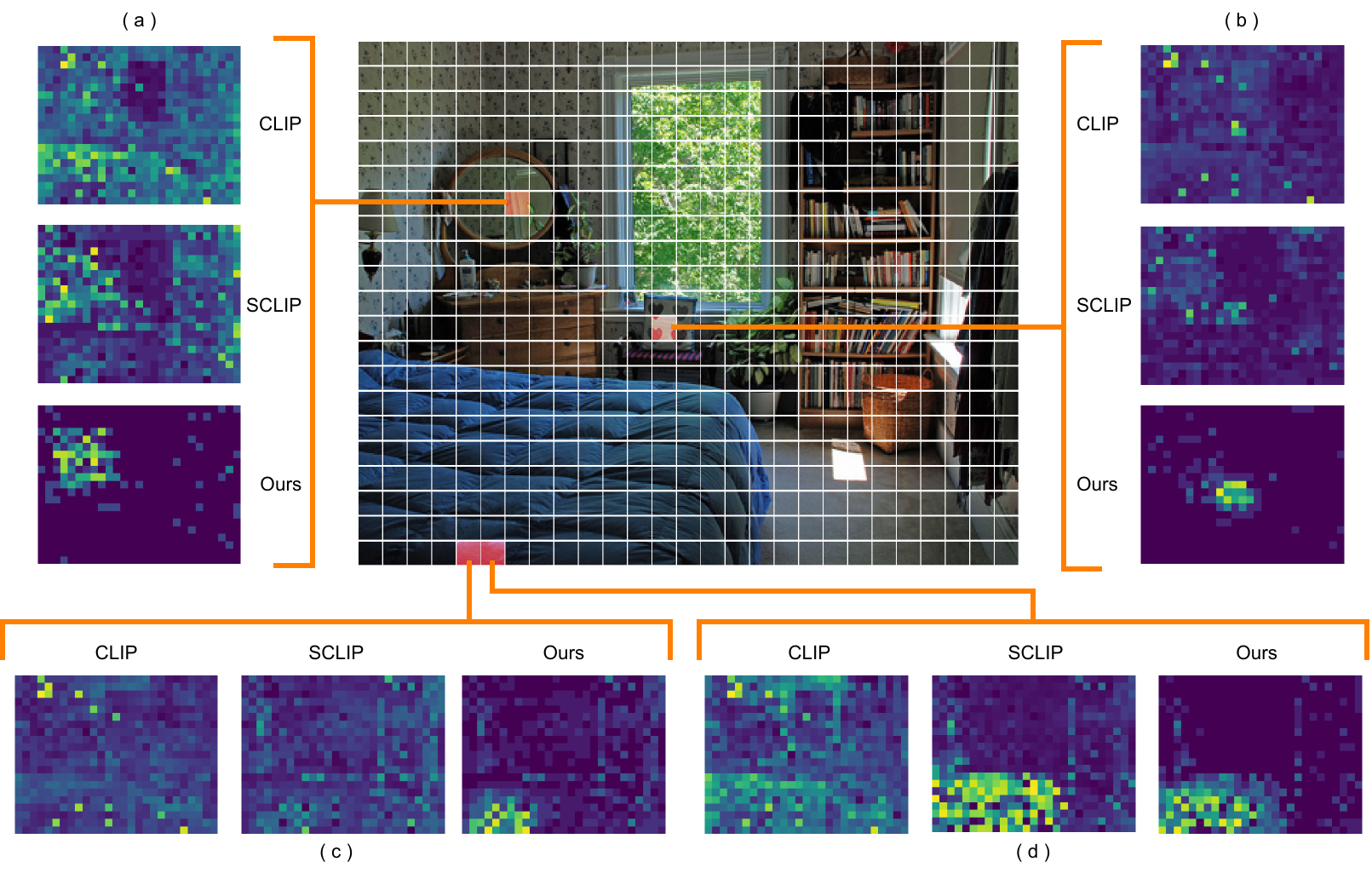}
    \caption{\textbf{Attention maps of the final visual encoder layer.} For the patches shaded in red (denoted with (a) to (d)), the final layer's attention maps are presented for CLIP \cite{clip}, SCLIP \cite{sclip}, and our method. 
    We have identified two problematic phenomena in the attention maps of CLIP and SCLIP, stemming from a lack of mechanisms to properly attend to patches' neighbourhoods.
    First, as depicted in (a) and (b), attention intensity is sometimes dispersed among distant patches, neglecting the vicinity of a patch.
    Additionally, adjacent or closely located patches sharing the same real-world category and even similar visual characteristics can have inconsistent attention maps. 
    For instance, while SCLIP generates a quality attention map for patch (d), its attention map for (c) is notably different and fails to focus on the desired object. 
    By explicitly promoting attention to neighbours, our method produces consistent attention maps across adjacent patches.}
    \label{fig:attn_maps}
\end{figure*}

In dense prediction tasks such as semantic segmentation, the crucial aspect of localization is often overlooked when employing vision transformers (ViTs) \cite{vit}. 
Notably, ViTs emphasize global representations, particularly through their \texttt{[CLS]} token, which leads to suboptimal dense prediction performance. 
Therefore, we advocate that devising novel and effective methodologies hinges on considering design choices that favour segmentation.
The recent concurrent work, SCLIP \cite{sclip}, shares a similar line of research, and investigates the inherent problems of CLIP in dense prediction, proposing minor adjustments over the baseline model. In particular, the authors identified that the baseline's poor performance is caused by a spatial misalignment of the patch representations, pointing to CLIP’s self-attention modules as the source of the problem. More concretely, they argue that CLIP learns spatial-invariant visual features, beneficial only in image-level tasks.
To address this limitation,
a new self-attention mechanism was introduced in \cite{sclip}, which re-organizes the spatial information. Nevertheless, despite its remarkable performance compared to CLIP, it does not guarantee semantic correlations across local tokens, as it does not have an explicit mechanism to attend to each token's neighbours and therefore ensure spatial consistency. As depicted in \cref{fig:attn_maps}, although SCLIP \cite{sclip} captures semantic context better than the standard attention in CLIP, it yields unstable attention maps on nearby tokens.

\paragraph{Contributions.}
Motivated by these limitations, in this work, we do due diligence in \textit{training-free} OVSS and propose a straightforward baseline with minimal modifications to vanilla CLIP. The implemented changes are motivated by the specificity of dense prediction problems, which have been mostly disregarded in existing methods. Specifically, we identify the potential limitations of CLIP in image segmentation and modify its visual encoder, particularly in the final layer, to enhance its localization capabilities.
To ensure spatial consistency, we propose a simple solution that encourages each patch to attend to its neighbours, generating consistent attention maps across adjacent patches. As demonstrated empirically through comprehensive experiments on 8 popular OVSS benchmarks, our strong baseline Neighbour-Aware CLIP (\ourmethod), achieves state-of-the-art performance without requiring additional data, auxiliary pre-trained networks, or extensive hyperparameter tuning. 

\section{Related work}
\subsection{Adapting large-scale vision-language models}

The field of deep learning is undergoing a learning paradigm shift with the emergence of vision-language models trained at a large scale. In particular, the Contrastive Language-Image Pre-Training (CLIP) method \cite{clip} has shown unprecedented success, mainly due to its remarkable
zero-shot and few-shot transfer capabilities on visual recognition tasks, notably in the context of classification. Fueled by its generalization and transfer capabilities, a realm of approaches have arisen to either improve its zero-shot performance by modifying its training on image-text pairs \cite{yang2022unified,wu2023medklip,fini2023improved}, or efficiently adapt to novel tasks with only a few labeled images \cite{zhang2021tip,gao2021clip,yu2023task,sung2022vl,hu2021lora,silva2023closer}. Nonetheless, the pre-training is performed at the image level, and therefore, typically, only the class token is trained to capture the global information, which limits the applicability of these approaches for dense prediction tasks. 

\subsection{Open-vocabulary semantic segmentation}

The outstanding transferability of CLIP in classification has surged to a rapidly growing literature in open-vocabulary semantic segmentation (OVSS) \cite{barsellotti2023enhancing,ghiasi2022scaling,liang2023open,xu2023open,xu2023side,xu2022simple,maskclip,reco,ovsegmentor,segclip,xu2023side,luo2024emergent}, which attempts to segment novel concepts in a given image without explicit supervision for them. These methods can be categorized into three main groups, depending on the level of auxiliary data required for adaptation: \textit{fully-supervised}, \textit{weakly-supervised}, and \textit{training-free}. \textit{Fully-supervised} OVSS approaches \cite{barsellotti2023enhancing,ghiasi2022scaling,liang2023open,xu2023open,xu2023side,xu2022simple,qin2023freeseg} adapt the pre-trained CLIP to semantic segmentation by exploiting a large labeled set containing pixel-wise annotations from a set of classes. 
The main idea of these approaches is that, by leveraging a large pixel-labeled dataset of an arbitrary limited set of categories, the model can learn to perform segmentation, while maintaining the remarkable transfer capability of CLIP. Nevertheless, the adaptation datasets employed in this family of approaches typically present a high class overlap with the testing images. Instead of requiring pixel-wise masks, \textit{weakly-supervised} OVSS resorts to additional datasets with image-level tags, typically in the form of image-text pairs \cite{chen2023exploring,cai2023mixreorg,zhang2024uncovering,wu2023clipself,groupvit,liu2022open,tcl}. A common strategy is to use these large image-text pair sets as supervision during adaptation, where the categories present in an image are included in the text. The alignment between the visual and textual information is typically performed via a contrastive loss \cite{groupvit}, similar to the pre-training of CLIP, which can be further enhanced by integrating additional learning strategies, such as online pixel clustering \cite{liu2022open} or multiple contrastive losses between the features from original images and those restored from corrupted versions \cite{cai2023mixreorg}, for example. Nevertheless, a substantially large dataset is still required for adaptation, and classes appearing in each image need to be known in advance, presenting impractical considerations for real-world problems. 

Given the points raised above, we focus in this work on \textit{training-free} OVSS where, ideally, access to additional data for adaptation is not allowed. Nevertheless, as stated earlier, many of the methods rather rely on leveraging auxiliary models pre-trained on large-scale datasets (\eg, ViT \cite{reco,wysoczanska2024clip,luo2024emergent} or stable diffusion \cite{barsellotti2024fossil,ovdiff,freesegdiff}), or containing multiple components whose hyperparameters must be tuned \cite{barsellotti2024fossil,luo2024emergent}. A different line of work attempts to improve the potential of CLIP in extracting dense visual features by modifying ViT's self-attention \cite{maskclip,bousselham2023grounding,clipsurgery}. This can be done, for instance, by directly using the value vectors \cite{maskclip}, or by introducing additional pathways computing self-self attentions in parallel to the encoder blocks to aggregate output of several layers using the residual connections \cite{clipsurgery,bousselham2023grounding}. A main difference with this line of work, however, is that we introduce explicit local spatial consistency in the self-attention, a concept that has been disregarded by the existing \textit{training-free} OVSS literature.

The concurrent approach, SCLIP \cite{sclip}, proposes a self-attention mechanism encouraging each token to attend to itself and the positions sharing similar information. In particular, the authors state that an attention map with maximum values on the diagonal leads to proper localization. Nevertheless, since query and key vectors are not normalized in SCLIP, there is no guarantee that the maximum values of the attention map fall on the diagonal. Hence, if an outlier patch has a key vector with a magnitude much larger than the others, patches attend to the outlier more than themselves. Furthermore, even though in a well-localized model each patch should attend to itself with a high intensity, localization is not limited to self, and the vicinity of each patch should be taken into account as well. This is particularly important in segmentation, where adjacent patches often represent the same class. As apparent in our qualitative analysis (\cref{fig:attn_maps}), despite SCLIP improving the spatial localization of attention maps over CLIP, it often generates contrasting attention maps for adjacent patches containing the same real-world object. This demonstrates that in this method, spatial localization in the neighbourhood of a patch does not typically go far beyond the patch itself. Moreover, although for a given patch, SCLIP attends to the patches that share similar query or key vectors, neighbouring patches do not necessarily do. 
We argue that only attending to patches that share similar vectors (and also self) is suboptimal, which is demonstrated empirically in \cref{fig:attn_maps}. Our proposed method can handle these limitations by adding an explicit mechanism that enforces patches to attend to their neighbours, imposing the local spatial consistency required for semantic segmentation.

\section{Preliminaries}

In this work, we address the task of open-vocabulary semantic segmentation in a \textit{training-free} scenario, \ie, with no supervision and without fine-tuning the parameters. Hence, for a given image $\XX_i \in \real^{\Omega_i}$, where $\Omega_i$ denotes its spatial domain $H \times W \times \channel$, and a set of arbitrary concepts described by natural language $\ttt_j \in \mathcal{T}$, our goal is to provide a segmentation mask of concepts present in the image.
In particular, an image $\XX_i$ is fed into CLIP's ViT, which produces a representation of each patch. In parallel, \mbox{$\ttt_j \; \forall j$} are fed into CLIP's text encoder and the resulting representations are compared to each patch's representation using cosine similarity, forming a probability distribution which is converted to a segmentation mask using an $\arg\max$ operation. 

\subsection{Background on CLIP} \label{sec:clip}

CLIP \cite{clip} employs a joint training approach, aligning visual and text modalities into the same feature space. Since most of CLIP-based OVSS methods utilize the vision transformer \cite{vit} version of CLIP, we focus on this architecture in what follows.
In this model, the visual encoder comprises $L$ sequential blocks, each processing $1 + hw$ tokens: the first one representing the \texttt{[CLS]} token, which captures global information, and the subsequent tokens each representing a patch. Thus, given an input image, it is initially partitioned into $hw$ non-overlapping patches ($h = H/P$ and $w = W/P$) each in $\real^{P \times P \times \channel}$, where $(P, P)$ denotes the resolution of each patch. Then, a linear transformation projects the sequence of $1 + hw$ tokens from $\channel$ channels to a $D$-dimensional space, and positional embeddings are added to create the input for the first encoder block. The components of this model are described below. For the sake of simplicity of our method's formulation in \cref{sec:new-self-attn}, we avoid flattening the 2D grid of patches and do not consider the \texttt{[CLS]} token in the formulations below.

\paragraph{Encoder block.} 
Each encoder block $l$ receives a sequence of tokens, $\ZZ^{(l-1)}$, from the preceding block and performs the following operations:
\begin{align}
    \phantom{,} \ZZ' &= \operatorname{LN}(\ZZ^{(l-1)}), \label{eq:encoder-l1} \\
    \phantom{,} \ZZ' &= \ZZ^{(l-1)} + \operatorname{SA}(\ZZ'), \label{eq:encoder-l2} \\
    \phantom{,} \ZZ^* &= \operatorname{LN}(\ZZ'), \label{eq:encoder-l3} \\
    \phantom{.} \ZZ^{(l)} &= \ZZ' + \operatorname{MLP}(\ZZ^*). \label{eq:encoder-l4}
\end{align}

\noindent In these equations, $\operatorname{LN}$, $\operatorname{SA}$ and $\operatorname{MLP}$ denote layer normalization, the self-attention module, and the feed-forward neural network, respectively. 
Please note that skip connections refer to the addition of $\ZZ^{(l-1)}$ and $\ZZ'$ in \cref{eq:encoder-l2,eq:encoder-l4}. Furthermore, the combination of \cref{eq:encoder-l1,eq:encoder-l2} is commonly referred to as the \textit{self-attention block}, whereas the combination of \cref{eq:encoder-l3,eq:encoder-l4} is called the \textit{feed-forward block}.

\paragraph{Self-attention module.} 
Within the self-attention module, the sequence of token representations undergoes a linear transformation $\mbf{W}^{qkv}$, yielding three sequences of $d$-dimensional vectors: query ($\qq$), key ($\kk$), and value ($\vv$). 
Subsequently, a similarity measure is calculated between all the tokens. More specifically, for a given patch at position $(i, j)$ the dot product of $\qq_{ij}$ and $\kk_{mn}$ is computed for all \mbox{$m \in \{1, 2, \dots, h\}$} and \mbox{$n \in \{1, 2, \dots, w\}$}. This measure is then scaled by $1 / \sqrt{d}$ and passed through a $\softmax$ operation to calculate a weighted sum of $\vv$ for the patch. Finally, the output is obtained via a projection using a linear transformation $\mbf{W}^{o}$. Thus, formally, operation $\operatorname{SA}$ for the patch at position $(i, j)$ can be described as:
\begin{align}
    \phantom{,} [\qq, \kk, \vv] &= \ZZ \mbf{W}^{qkv}, \label{eq:orig-sa-l1} \\
    \phantom{,} \text{sim}_{ij} &= \frac{\kk \qq_{ij}}{\sqrt{d}}, \label{eq:orig-sa-l2} \\
    \phantom{,} A_{ij} &= \softmax\left(\text{sim}_{ij}\right) \vv, \label{eq:orig-sa-l3} \\
    \phantom{.} \operatorname{SA}(\ZZ)_{ij} &= A_{ij} \mbf{W}^{o}. \label{eq:orig-sa-l4}
\end{align}

\noindent Hereafter, we will refer to $\softmax\left(\text{sim}_{ij}\right)$ as the \textit{attention map} of point $(i, j)$, and $\text{sim}_{ij}$ as its corresponding logits. Note that while in practice a multi-head version of the self-attention is used, for simplicity, we have presented equations considering a single head.

\subsection{Limitations of CLIP for image segmentation} \label{sec:clip-limitation}

As previously stated, localization holds a pivotal role in segmentation tasks, yet it is often overlooked in ViTs \cite{vit}. As described in \cite{vit}, unlike CNNs, most of the operations in ViTs are global and the locality of patches is mostly not taken into account. Within ViT, and consequently, in CLIP's visual encoder, positional information is integrated into the network through 1D positional embeddings added to the input at the first block. Recent evidence points to the possible suboptimal performance of 1D embeddings in image data, and thus alternative mechanisms are sought after \cite{liu2021swin,chu2023conditional}.
Furthermore, positional embeddings utilization is restricted only to the first encoder block, potentially diminishing their relevance in subsequent layers. This oversight could be detrimental in dense prediction tasks, where patches contain important contextual information, and such information remains valuable throughout the network's depth. Please note that simply incorporating the pre-trained positional embeddings at all the layers is not feasible in the \textit{training-free} scenario since such additions would change the distribution of the data that each layer expects and requires fine-tuning.
In light of the arguments presented, we assert that explicitly attending to the neighbourhood of each patch is imperative within the context of segmentation.
Moreover, recent studies have noted that the final encoder block in CLIP's ViT disrupts spatial information, hindering dense prediction tasks \cite{maskclip}.
Indeed, CLIP's visual encoder is trained to emphasize the \texttt{[CLS]} token's embedding (global embedding), while the outputs at other locations (\ie, embeddings of patches) are not optimally structured for tasks such as semantic segmentation.

\section{Neighbour-Aware CLIP} \label{sec:our-method}

The pre-training procedure of CLIP \cite{clip} encourages its ViT \cite{vit} to learn representations tailored to image-level tasks, thereby compromising its efficacy in dense prediction problems. Given the inherent differences between such tasks and segmentation, coupled with the fact that the output of the ViT for patch tokens has not been explicitly trained during CLIP's pre-training, it struggles to generalize effectively to pixel-wise prediction scenarios. This underscores the necessity for targeted adjustments in the original CLIP model to accommodate the nuances of semantic segmentation. Our study delves into these specific components that potentially hinder CLIP's segmentation performance and proposes minimal alterations to its overall framework, without changing any of the network's parameters. The precise modifications are detailed in this section.

\subsection{Introducing spatial consistency} \label{sec:new-self-attn}

In \cref{sec:clip-limitation}, we underscored the importance of explicitly attending to the locality of each patch and highlighted the inadequacy of vanilla ViT's positional embeddings in achieving this. In this section, we introduce our simple method for enforcing explicit spatial attention to each patch's neighbourhood. In particular, we augment the attention map information with an unnormalized multivariate (in our case, $2$-dimensional) Gaussian kernel, which can be defined as:
\begin{equation}
    \phantom{.} \phi(\xx; \mmu, \boldsymbol{\Sigma}) = \exp \left(-{\frac {1}{2}}(\xx - \mmu)^\T \boldsymbol{\Sigma}^{-1}(\xx - \boldsymbol{\mu})\right).
\end{equation}

\noindent Assuming $\boldsymbol{\Sigma} = \sigma^2 I$, we can rewrite the kernel as
\begin{equation}
\label{eq:gaussian_func}
    \phantom{.} \phi(\xx; \mmu, \sigma) = \exp \left(- \frac {||\xx - \mmu||^2}{2 \sigma^2}\right),
\end{equation}

\noindent which is maximized when $\xx = \mmu$ and decreases as the Euclidean distance of $\xx$ to $\mmu$ increases.
We now define a function $\omega((i, j))$ that, given a coordinate as input, discretizes $\phi$ and outputs a matrix of size $h \times w$:
\begin{equation}
\label{eq:gaussian_window}
    \begin{gathered}
        \omega((i, j); \sigma)_{mn} = \phi((m, n); {(i, j)}, \sigma), \\
        \phantom{,} \forall m \in \{1, 2, \dots, h\}, \; \forall n \in \{1, 2, \dots, w\},
    \end{gathered}
\end{equation}
with the maximum value at $(i, j)$, gradually decreasing with distance from $(i, j)$ (see $\omega$ in \cref{fig:schematic} for elaboration).

As a corner case, let us assume that the attention map's logits have been explicitly set to $0$ and we set
\begin{equation} \label{eq:og}
    \phantom{,} A_{ij} = \softmax\left(\omega((i, j), \sigma)\right) \vv,
\end{equation}

\noindent so that the attention will be purely paid to the neighbourhood of patches. As we show in our empirical validation (termed \og in \cref{sec:abl}), by just doing so we can observe a leap in performance compared to CLIP.

Examining \cref{eq:og}, we observe that the patch information is not utilized as $\omega$ replaces $\text{sim}_{ij}$ in \cref{eq:orig-sa-l3}, leading to image-independent attention. This prompts us to move beyond the corner case and include the similarity information:
\begin{gather}
    \phantom{,} \widetilde{A}_{ij} = \softmax\left(\text{sim}_{ij} + \omega((i, j); \sigma) \right) \vv, \label{eq:new-self-attention} \\
    \phantom{.} \oursa(\ZZ)_{ij} = \widetilde{A}_{ij} \mbf{W}^{o}. 
\end{gather}

\noindent By adding the Gaussian window to the logits of the attention map for patch $(i, j)$, the attention increases not only to $\vv_{ij}$, which has been proven fruitful in semantic segmentation \cite{maskclip,sclip}, but also to the value vectors of the patches in the vicinity of $(i, j)$, and therefore locality is introduced to the model.

\subsection{Measure of similarity} \label{sec:kk-sim}
Transformers \cite{transformer} are complex deep neural networks and providing a definitive explanation of their inner workings remains challenging. However, from an intuitive standpoint, we can offer the following interpretations for query, key, and value vectors. The query vector denotes
what a patch \emph{looks for}; the key vector signifies what it \emph{represents}; and the value vector displays what it \emph{has to offer}. Guided by these descriptions, we deviate from the standard similarity measure ($\qq \kk^\T$) in our self-attention module.
This deviation is motivated by the discrepancy between what we want the model to look for (\ie, accurate patch-level predictions) and what it had been trained to look for during pre-training.
In semantic segmentation, we need to focus on the nature of each patch and this naturally leads us to shift our attention toward using the key vectors.
Thus, we opted to use $\kk \kk^\T$ scores in our similarity measure instead, resulting in
$\text{sim}_{ij} = {\kk \kk_{ij}} / {\sqrt{d}}$.
By doing so, patches that represent similar information (as portrayed by their key vectors) attend to each other's value vectors with high intensity. This simple change in perspective brings considerable performance improvement compared to CLIP, as shown in our empirical validation (termed \kksim in \cref{sec:abl}).

\begin{figure}[t]
    \centering
    \includegraphics[width=\columnwidth]{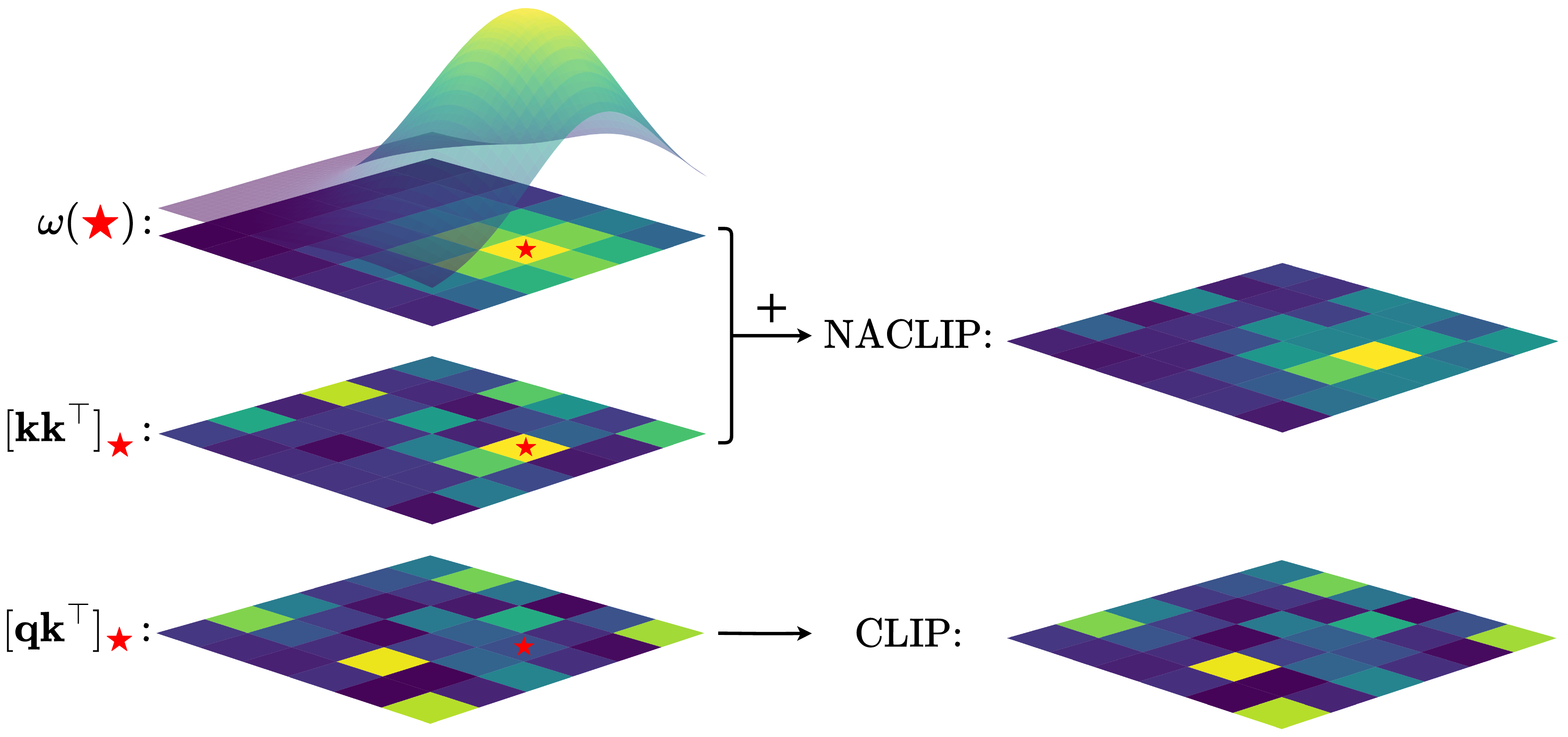}
    \caption{\textbf{Schematic figure depicting the mechanism to form attention maps.}
    Maps are shown for the patch located at $\color{red} \bigstar$, 
    $\omega({\color{red} \bigstar})$ denotes a discretized Gaussian kernel centered at $\color{red} \bigstar$ (see \cref{eq:gaussian_window}), and
    $[{\mathbf{x} \mathbf{y}^{\top}}]_{\color{red} \bigstar}$ indexes $\mathbf{x} \mathbf{y}^{\top}$ on patch $\color{red} \bigstar$.
    CLIP does not ensure high attention to the patch itself and the neighbouring patches, while \ourmethod does. Scaling and $\mathrm{softmax}$ operations are omitted for demonstration simplicity.}
    \label{fig:schematic}
\end{figure}

Considering the changes proposed up until now, we rewrite \cref{eq:new-self-attention} as (see \cref{fig:schematic} for more insight):
\begin{equation} \label{eq:final-self-attention}
    \phantom{.} \widetilde{A}_{ij} = \softmax\left(\frac{\kk \kk_{ij}}{\sqrt{d}} + \omega((i, j); \sigma) \right) \vv.
\end{equation}

\subsection{Eliminating image-level specialized units} \label{sec:reducing}

As outlined in \cref{sec:clip-limitation}, the final encoder block of CLIP's vision transformer undermines the network's efficacy for dense prediction tasks \cite{maskclip}.
Therefore, we have chosen to eliminate specific modules from the final encoder, rendering CLIP more suitable for semantic segmentation. Specifically, we have removed the feed-forward block from this encoder, as its parameters are tailored for image-level tasks rather than dense prediction.
Additionally, with the alteration of the self-attention operation to incorporate locality and the removal of the feed-forward block, the inclusion of a skip connection becomes impractical. This is because it puts greater importance on the output of the previous encoder block, thereby diminishing the significance of the output of our self-attention module ($\oursa$). 
Considering these modifications, the final visual encoder block in our approach simplifies the operations described in \cref{sec:clip} to
\begin{equation}
    \phantom{,} \ZZ^{(L)} = \oursa\left(\operatorname{LN}\left(\ZZ^{\left(L-1\right)}\right)\right),
\end{equation}

\noindent in which $L$ denotes the index of the last encoder block. We refer to this structure as the \textit{Reduced} architecture.

\section{Experiments}

\begin{figure*}[tb]
    \centering
    \includegraphics[width=0.9\textwidth]{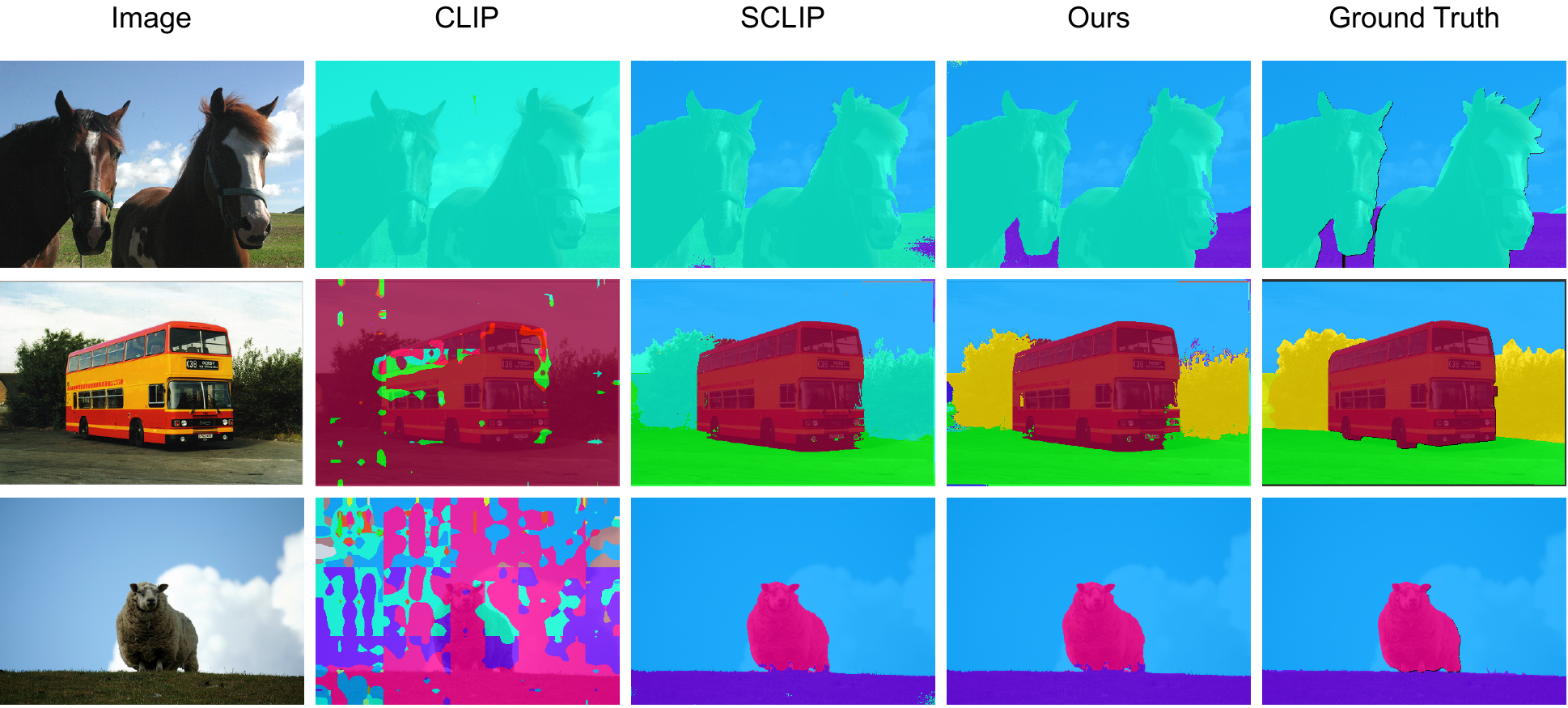}
    \caption{\textbf{Qualitative results (segmentation maps) on PASCAL Context~(59)~\cite{pascalcontext}} for CLIP \cite{clip}, SCLIP \cite{sclip}, and our method.}
    \label{fig:qualitative}
\end{figure*}

\begin{table*}[tbh]
    \centering
    \caption{
    \textbf{Quantitative evaluation on 6 datasets and 2 of their variants.} The first 3 benchmarks (V21, PC60, and C-Obj) contain a background category, while the subsequent ones do not. The \textit{Fair} column indicates whether a method's comparison to ours is equitable, \ie, conducted without leveraging additional knowledge. The \textit{Post.} column shows whether an approach contains a post-processing step for mask refinement.
    For elucidation on abbreviated benchmark names, please refer to \cref{sec:exp_setup}.
    }    
        \begin{tabular}{l@{\hskip 5px}lccccccccccc}
        \toprule
        \textbf{Method} && Fair & Post. & V21 & PC60 & C-Obj & V20 & City & PC59 & ADE & C-Stf & Avg \\
        \midrule
        CLIP~\cite{clip}                    & {\scriptsize ICML'21}       & \cmark & \xmark & 18.6 &  7.8 &  6.5          & 49.1          &  6.7 & 11.2 &  3.2 &  5.7 & 13.6 \\ 
        MaskCLIP~\cite{maskclip}            & {\scriptsize ECCV'22}       & \cmark & \xmark & 43.4 & 23.2 & 20.6          & 74.9          & 24.9 & 26.4 & 11.9 & 16.7 & 30.3 \\
        GroupViT~\cite{groupvit}            & {\scriptsize CVPR'22}       & \xmark & \xmark & 52.3 & 18.7 & 27.5          & 79.7          & 18.5 & 23.4 & 10.4 & 15.3 & 30.7 \\
        CLIP Surgery~\cite{clipsurgery}     & {\scriptsize arXiv'23}      & \cmark & \xmark & 41.2 & 30.5 & -             & -             & 31.4 & -    & 12.9 & 21.9 & -    \\
        CLIP-DIY~\cite{wysoczanska2024clip} & {\scriptsize WACV'24}       & \xmark & \xmark & 59.0 & -    & 30.4          & -             & -    & -    & -    & -    & -    \\
        GEM~\cite{bousselham2023grounding}  & {\scriptsize CVPR'24}       & \cmark & \xmark & 46.2 & -    & -             & -             & -    & 32.6 & 15.7 & -    & -    \\
        SCLIP~\cite{sclip}                  & {\scriptsize ECCV'24}      & \cmark & \xmark & \textbf{59.1} & 30.4 & 30.5 & \textbf{80.4} & 32.2  & 34.2 & 16.1 & 22.4 & 38.2 \\
        \rowcolor{mygray} \textbf{\ourmethod} & {\scriptsize Ours} & \cmark & \xmark & 58.9 & \textbf{32.2} & \textbf{33.2} & 79.7 & \textbf{35.5} & \textbf{35.2} & \textbf{17.4} & \textbf{23.3} & \textbf{39.4} \\ 
        \midrule
        ReCo~\cite{reco}                    & {\scriptsize NeurIPS'22}    & \xmark & \cmark & 25.1 & 19.9 & 15.7          & 57.7          & 21.6 & 22.3 & 11.2 & 14.8 & 23.5 \\
        TCL~\cite{tcl}                      & {\scriptsize CVPR'23}       & \xmark & \cmark & 55.0 & 30.4 & 31.6          & 83.2          & 24.3 & 33.9 & 17.1 & 22.4 & 37.2 \\ 
        FreeSeg-Diff~\cite{freesegdiff}     & {\scriptsize arXiv'24}      & \xmark & \cmark & 53.3 & -    & 31.0          & -             & -    & -    & -    & -    & -    \\ 
        FOSSIL~\cite{barsellotti2024fossil} & {\scriptsize WACV'24}       & \xmark & \cmark & -    & -    & -             & -             & 23.2 & 35.8 & 18.8 & 24.8 & -    \\
        PnP-OVSS~\cite{luo2024emergent}     & {\scriptsize CVPR'24}       & \xmark & \cmark & 51.3 & -    & \textbf{36.2} & -             & -    & 28.0 & 14.2 & 17.9 & -    \\
        SCLIP~\cite{sclip}                  & {\scriptsize ECCV'24}      & \cmark & \cmark & 61.7 & 31.5 & 32.1          & \textbf{83.5} & 34.1 & 36.1 & 17.8 & 23.9 & 40.1 \\
        \rowcolor{mygray} \textbf{\ourmethod} & {\scriptsize Ours} & \cmark & \cmark & \textbf{64.1} & \textbf{35.0} & \textbf{36.2} & 83.0 & \textbf{38.3} & \textbf{38.4} & \textbf{19.1} & \textbf{25.7} & \textbf{42.5} \\  
        \bottomrule
        \end{tabular}
    \label{tab:main}
\end{table*}

\begin{table}[tbh]
    \centering
    \caption{\textbf{Evaluation using different CLIP-ViT backbones.}}
    \resizebox{\columnwidth}{!}{
        \begin{tabular}{@{}l@{\hskip 5px}|@{\hskip 5px}c@{\hskip 5px}c@{\hskip 5px}c@{\hskip 5px}|@{\hskip 5px}c@{\hskip 5px}c@{\hskip 5px}c@{\hskip 5px}|@{\hskip 5px}c@{\hskip 5px}c@{\hskip 5px}c@{}}
        \toprule
        & \multicolumn{3}{c|@{\hskip 5px}}{\textbf{ViT-B/16}} & \multicolumn{3}{c|@{\hskip 5px}}{\textbf{ViT-B/32}} & \multicolumn{3}{c}{\textbf{ViT-L/14}} \\
        Method & V21 & PC59 & Avg. & V21 & PC59 & Avg. & V21 & PC59 & Avg. \\
        \midrule
        SCLIP~\cite{sclip} & 61.7 & 36.1 & 48.9 & \textbf{54.8} & 30.2 & 42.5 & 45.4 & 27.4 & 36.4\\
        GEM~\cite{bousselham2023grounding} & 46.2 & 32.6 & 39.4 & 40.5 & 27.0 & 33.8 & 44.6 & 28.6 & 36.6 \\
        \rowcolor{mygray}
        \textbf{\ourmethod} & \textbf{64.1} & \textbf{38.4} & \textbf{51.3} & \textbf{54.8} & \textbf{34.9} & \textbf{44.9} & \textbf{57.9} & \textbf{36.4} & \textbf{47.2}\\
        \bottomrule
        \end{tabular}
    }
    \label{tab:backbone}
\end{table}

\subsection{Experimental setup} \label{sec:exp_setup}

\paragraph{Datasets.}
We evaluate our method on the following segmentation benchmarks, whose names are abbreviated (in parentheses) to conserve table space:
PASCAL VOC 2012 (V21) \cite{voc12},
ADE20K-150 (ADE) \cite{ade20k},
PASCAL Context (PC60) \cite{pascalcontext},
COCO-Stuff (C-Stf) \cite{coco},
Cityscapes (City) \cite{cityscapes},
COCO-Object (C-Obj) \cite{actualcoco}.
Additionally, alongside the original benchmarks on these datasets, we follow \cite{sclip} and evaluate on variants of PASCAL VOC 2012 (V20) and PASCAL Context (PC59) in which the background class is removed from the evaluation. 
Furthermore, input images are resized to have a shorter side of 336 (560 for Cityscapes \cite{cityscapes}, because of high-resolution images), and following prior works \cite{sclip,barsellotti2024fossil,clipsurgery,tcl,groupvit}, a slide inference is carried out with a $224 \times 224$ window and a stride of 112.

\paragraph{Baselines.} 
We compare our method to a set of relevant works in OVSS, including:
MaskCLIP~\cite{maskclip}, ReCo~\cite{reco}, CLIP Surgery~\cite{clipsurgery}, SCLIP~\cite{sclip}, GEM~\cite{bousselham2023grounding}, CLIP-DIY~\cite{wysoczanska2024clip}, FreeSeg-Diff~\cite{freesegdiff}, FOSSIL~\cite{barsellotti2024fossil}, and PnP-OVSS~\cite{luo2024emergent}. 
We also include a few influential weakly supervised OVSS methods, such as GroupViT \cite{groupvit} and TCL \cite{tcl} in our comparison. 
Furthermore, since vanilla CLIP \cite{clip} can be adapted for semantic segmentation, we include it as a baseline in our comparison tables as well.

\paragraph{Implementation details.}
In our experiments, we employ pre-trained CLIP-ViT \cite{clip}. Unless mentioned otherwise, we use the ViT-B/16 backbone ($16 \times 16$ patch size) comprising 12 visual encoder blocks.
Since our approach operates in a \textit{training-free} manner, we exclusively utilize the frozen CLIP model without any optimization. 
The standard deviation of the Gaussian kernel, $\sigma$ in \cref{eq:gaussian_func}, is set to 5, a choice further elaborated upon in \cref{app:sec:gaussian}.
OVSS methods often incorporate a mask refinement step \cite{barsellotti2024fossil,sclip,tcl,reco,luo2024emergent,freesegdiff}, such as DenseCRF \cite{densecrf} or pixel-adaptive mask refinement (PAMR) \cite{pamr}. For our method, we opt for PAMR as it is lighter and more efficient. We also report results without this refinement step. We resort to mIoU for the evaluation metric across all experiments. 

\subsection{Main results}

\paragraph{Comparison to \textit{training-free} OVSS methods.}
In \cref{tab:main}, we evaluate our proposed method against baselines mentioned in \cref{sec:exp_setup}.
Although most methods are presented as \textit{training-free} OVSS, it is important to note that they vary in underlying characteristics. Specifically, within the subset of methods that refrain from fine-tuning, some leverage auxiliary pre-trained models \cite{reco,barsellotti2024fossil,wysoczanska2024clip,luo2024emergent,freesegdiff}. Given our primary focus on achieving \textit{training-free} OVSS exclusively through a frozen CLIP model, comparing our method to those utilizing additional weights and data, while informative, may not be entirely equitable. Hence, we explicitly denote fair comparisons in \cref{tab:main}.
Analyzing the results, our approach outperforms state-of-the-art \textit{training-free} OVSS methods in 7 out of 8 benchmarks. This signifies the effectiveness of our proposed architecture and formulation of CLIP's visual encoder. The improvement trend is also noticeable among methods that opted not to perform mask refinement steps compared to our method without post-processing.
Note that results for newer methods \cite{barsellotti2024fossil,wysoczanska2024clip,bousselham2023grounding,clipsurgery,luo2024emergent,freesegdiff} are extracted from their respective manuscripts, hence might not include experiments across all benchmarks, whereas other results have been derived from \cite{sclip}.

\paragraph{Robustness to visual backbones.}
\Cref{tab:backbone} reports the segmentation results of \ourmethod using different CLIP-ViT backbones, namely ViT-B/16 (default), ViT-B/32, and ViT-L/14. Compared to concurrent approaches \cite{sclip,bousselham2023grounding}, our method is more robust to the backbone choice. For example, SCLIP \cite{sclip} experiences a significant performance drop of over 12\% on average when using ViT-L/14 instead of the default backbone, whereas the degradation is about 4\% in our case. Moreover, our method outperforms GEM \cite{bousselham2023grounding} by over 10\% across all settings, making \ourmethod a more robust solution compared to \cite{sclip,bousselham2023grounding}.

\paragraph{Visual examples.}
\Cref{fig:qualitative} presents several segmentation maps generated by our method on the PASCAL Context~(59) dataset \cite{pascalcontext}. These visualizations demonstrate \ourmethod's significant performance enhancement over CLIP. 
Furthermore, SCLIP seems to find difficulties in identifying the limits of objects properly, sometimes confusing nearby concepts (\eg, the first row). Besides, while SCLIP tends to focus mainly on similar patches without explicitly considering the surrounding context, our method maintains a local contextual understanding. This is evident in the second row (the bus image) where SCLIP misclassifies trees, likely due to its limited attention to nearby objects such as the road. More examples are available in the \cref{app:visual}.

\subsection{Ablation study} \label{sec:abl}

\paragraph{Effect of altering the self-attention module.} 
In \cref{sec:new-self-attn}, we discussed the rationale behind introducing spatial consistency to attention maps and described our approach. Besides, we defined the \og (\textit{\ogshort} in the table) setting, where patches exclusively attend to their neighbours, disregarding similarities.
Furthermore, we defined the \kksim (\textit{\kksimshort} in the table) setting in \cref{sec:kk-sim}, in which $\kk \kk^\T$ is used as the similarity measure. 
Results from the comparative analysis (\cref{tab:abl-attn}) between \mbox{\textit{\ogshort}} and \textit{Vanilla}, as well as between \textit{\kksimshort} and \textit{Vanilla} underscore the efficacy of the modifications proposed in \cref{sec:new-self-attn,sec:kk-sim}, respectively.

\begin{table}[tb]
    \centering
    \caption{\textbf{Ablation on the effect of spatial consistency and similarity measure.} \textit{Vanilla} refers to CLIP's self-attention module, $\widetilde{A}$ has been defined in \cref{eq:final-self-attention}, and the rest of the settings have been mentioned in \cref{sec:abl}.
    Throughout this experiment, we adhere to the default CLIP encoder architecture, retaining all architectural elements of the encoder block.}
    \resizebox{\columnwidth}{!}{
        \begin{tabular}{l@{\hskip 10px}c@{\hskip 5px}c@{\hskip 5px}c@{\hskip 5px}c@{\hskip 5px}c@{\hskip 5px}c@{\hskip 5px}c@{\hskip 5px}c@{\hskip 5px}c}
        \toprule
        \textbf{Atten.} & V21 & PC60 & C-Obj & V20 & City & PC59 & ADE & C-Stf & Avg \\
        \midrule
        Vanilla          & 18.6 &  7.8 &  6.5 & 49.1 &  6.7 & 11.2 &  3.2 &  5.7 & 13.6 \\
        \ogshort         & 38.0 & 17.0 & 16.1 & 65.9 & 21.4 & 23.9 & 10.2 & 14.6 & 25.9 \\
        \kksimshort      & 36.0 & 15.6 & 10.9 & 64.9 & 24.8 & 26.2 &  9.6 & 15.1 & 25.4 \\
        $\widetilde{A}$  & 40.2 & 17.4 & 13.9 & 68.2 & 28.1 & 27.9 & 11.2 & 16.5 & 27.9 \\
        \bottomrule
        \end{tabular}
    }
    \label{tab:abl-attn}
\end{table}

\paragraph{Impact of architectural reduction.}
Here, we explore the effects of architectural modifications made to the final block of CLIP's visual encoder, as outlined in \cref{sec:reducing}. Specifically, we investigate the efficacy of utilizing the self-attention module's output as the output of the final encoder block. As showcased in \cref{tab:abl-arch}, the reduction in architecture and the removal of previously described operations yield substantial benefits in semantic segmentation tasks.

\begin{table}[tb]
    \centering
    \caption{\textbf{Investigating architectural reduction's impact.} We compare CLIP's default encoder architecture, denoted as \textit{Vanilla}, with the \textit{Reduced} setting described in \cref{sec:reducing}, where the self-attention module's output directly serves as the final encoder block's output. 
    Throughout this experiment, we utilize CLIP's default self-attention module.}
    \resizebox{\columnwidth}{!}{
        \begin{tabular}{l@{\hskip 10px}c@{\hskip 5px}c@{\hskip 5px}c@{\hskip 5px}c@{\hskip 5px}c@{\hskip 5px}c@{\hskip 5px}c@{\hskip 5px}c@{\hskip 5px}c}
        \toprule
        \textbf{Arch.} & V21 & PC60 & C-Obj & V20 & City & PC59 & ADE & C-Stf & Avg \\
        \midrule
        Vanilla        & 18.6 &  7.8 &  6.5 & 49.1 &  6.7 & 11.2 &  3.2 &  5.7 & 13.6 \\
        Reduced        & 37.5 & 22.3 & 23.2 & 81.4 & 20.2 & 24.9 & 11.6 & 16.7 & 29.7 \\
        \bottomrule
        \end{tabular}
    }
    \label{tab:abl-arch}
\end{table}

\section{Conclusion}

Drawing from CLIP's remarkable zero-shot generalizability, the paradigm of open-vocabulary semantic segmentation leveraging this model has gained significant traction, emerging as an appealing alternative to circumvent the limitations of traditional close-set supervised training.
In this work, we have explored the inherent weaknesses of CLIP for dense prediction and proposed simple and minimal modifications that accommodate this powerful model to the restrictive \textit{training-free} OVSS scenario. In addition to removing components of CLIP's visual encoder that hamper its localization capabilities, we have integrated a simple mechanism that explicitly encourages local consistency in the self-attention maps, which has been unexplored in the existing works. Extensive experiments on popular OVSS benchmarks showcase the superiority of our approach
over other existing OVSS methods, some of which resort to impractical or unrealistic choices, such as leveraging auxiliary models trained on additional large datasets or relying on validation sets for hyperparameter tuning. 
While yielding state-of-the-art performance on 7 out of 8 benchmarks, \ourmethod does not require access to either labeled or unlabeled data, becoming a suitable solution for real-world applications.

\paragraph{Limitations.}
In CLIP's pre-training, only the output at the \texttt{[CLS]} token's position directly influences optimization \cite{clip}. 
Although this token plays a primary role in the original CLIP's encoders, our attempts to extract information useful for segmentation from it were unsuccessful, having almost no effect on performance. We argue that this failure can be attributed to the divergent nature of CLIP's pre-training and the specific requirements of dense prediction problems, in which objects of interest must be recognized with their local positioning. 
Despite this, we acknowledge that the \texttt{[CLS]} token has proven effective in many image-level tasks and may contain information transferable to dense prediction. Therefore, we believe the use of this token in dense prediction requires further investigation.

\section*{Acknowledgments}

\noindent This work was funded by the Natural Sciences and Engineering Research Council of Canada (NSERC). We also thank Calcul Quebec and Compute Canada. 

{\small
\bibliographystyle{ieee_fullname}
\bibliography{main}
}

\clearpage
\appendix

\twocolumn[
    \begin{center}
        \section*{\Large 
            Supplementary Material for \\
            Pay Attention to Your Neighbours: \\
            Training-Free Open-Vocabulary Semantic Segmentation \\
            \hfill \\ \hfill \\
        }
    \end{center}
]

\section{Gaussian kernel's standard deviation} \label{app:sec:gaussian}

In a realistic \textit{training-free} open-vocabulary scenario, where additional data access is restricted, there should be no validation set available for hyperparameter tuning. Consequently, it is crucial for \textit{training-free} methods to operate effectively without such procedures. In our approach, we introduce a hyperparameter denoted as $\sigma$, representing the standard deviation of the Gaussian kernel used in \cref{eq:gaussian_func}, which we set to 5 in our experiments. In this section, we detail the heuristics guiding this choice, enabling us to determine this value without the need for fine-tuning.

For a patch located at $\mmu$, the Gaussian kernel increases its attention logits by $1$ at $\mmu$ and by lesser values at neighbouring patch locations.
Our choice of $\sigma$ is based on the number of neighbouring patches whose attention logits are modified by more than a threshold $\tau$. To achieve this, we express:
\begin{align}
    \phi(\xx; \mmu, \sigma) = \exp \left(- \frac {||\xx - \mmu||^2}{2 \sigma^2}\right) &> \tau \\
    \Rightarrow \; \frac {||\xx - \mmu||^2}{2 \sigma^2} &< - \ln \tau \\
    \Rightarrow \; \phantom{.} ||\xx - \mmu||^2 &< - 2 \sigma^2 \ln \tau \\
    \Rightarrow \; \phantom{.} ||\xx - \mmu||^2 &< \left( \sigma \sqrt{-2 \ln \tau} \right)^2. \label{app:eq:std-circle}
\end{align}

\begin{figure}[htb]
    \centering
    \includegraphics[width=0.9\columnwidth]{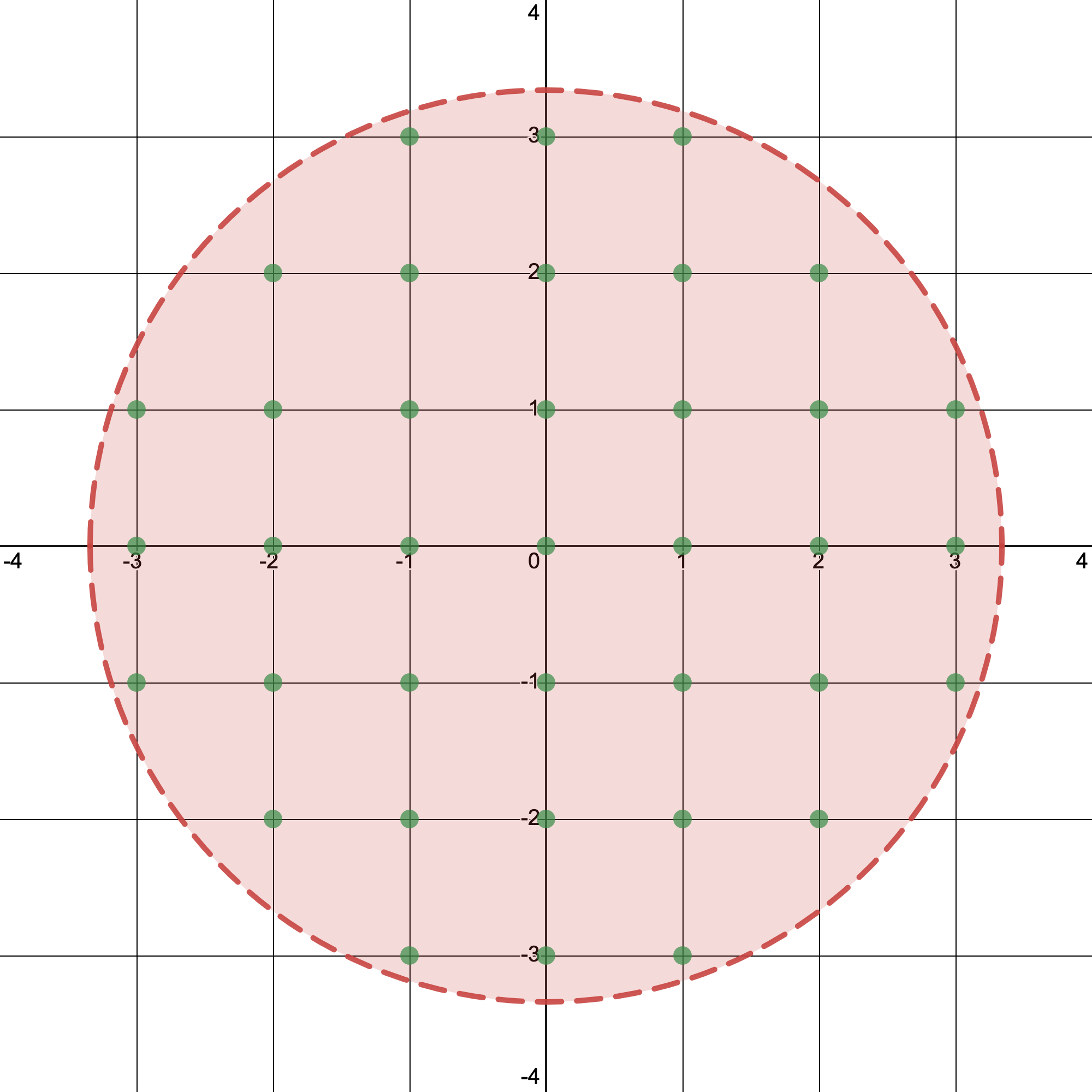}
    \caption{\textbf{Illustrative example of \cref{app:eq:std-circle}.} The attention logits of the center point to the points within the depicted circle are increased by at least $\tau$. Example generated for $\sigma = 5$ and $\tau = 0.8$.}
    \label{app:fig:std-circle}
\end{figure}

\noindent Considering \cref{app:eq:std-circle}, neighbouring patches for which
$\mmu$'s attention logits are increased by at least $\tau$ are positioned within of a circle centered on $\mmu$ with a radius of $\sigma \sqrt{-2 \ln \tau}$. For instance, with $\sigma = 5$, patch $\mmu$'s attention to 37 patches gets a logit increase of at least $0.8$ as illustrated in \cref{app:fig:std-circle}. 

\noindent \Cref{app:tab:std} displays the value of this heuristic measure for \mbox{$\sigma \in \{1, 2, \dots, 10\}$} and \mbox{$\tau \in \{0.7, 0.8, 0.9\}$}.
Besides, CLIP~\cite{clip} has been trained on $224 \times 224$ pixel images, meaning the ViT-B/16 backbone operates on $14 \times 14$ patches for each image. Based on this fact and considering the values provided in \cref{app:tab:std}, we opted for $\sigma = 5$ in our experiments to maintain a balance, \ie, to have neither too small nor too large field of attention. 
It is worth noting that $\tau$ is defined solely for the purpose of the described heuristic and does not play a role in our approach. In other words, there is no $\tau$ value to fine-tune in our approach.

\begin{table}[tbh]
    \centering
    \caption{\textbf{Proposed heuristic measure for determining $\sigma$ value.} For 3 values of $\tau$ and 10 values of $\sigma$, the table provides the number of patches that patch $\mmu$'s attention logits to them is increased by more than $\tau$.
    It is important to note that these values are derived based on an infinite grid of patches, while in practice, these numbers could be less depending on the window size and $\mmu$'s position.}
    \begin{tabular}{@{\hskip 10px}c@{\hskip 10px}|@{\hskip 10px}c@{\hskip 15px}c@{\hskip 15px}c}
        \toprule
        $\sigma$ & $\tau = 0.7$ & $\tau = 0.8$ & $\tau = 0.9$ \\
        \midrule
        1 & 1 & 1 & 1 \\
        2 & 9 & 5 & 1 \\
        3 & 21 & 13 & 5 \\
        4 & 37 & 21 & 9 \\
        5 & 57 & 37 & 21 \\
        6 & 81 & 49 & 21 \\
        7 & 109 & 69 & 37 \\
        8 & 145 & 89 & 45 \\
        9 & 177 & 113 & 57 \\
        10 & 221 & 137 & 69 \\
        \bottomrule
    \end{tabular}
    \label{app:tab:std}
\end{table}

\begin{figure*}[htb]
    \centering
    \includegraphics[width=0.6\textwidth]{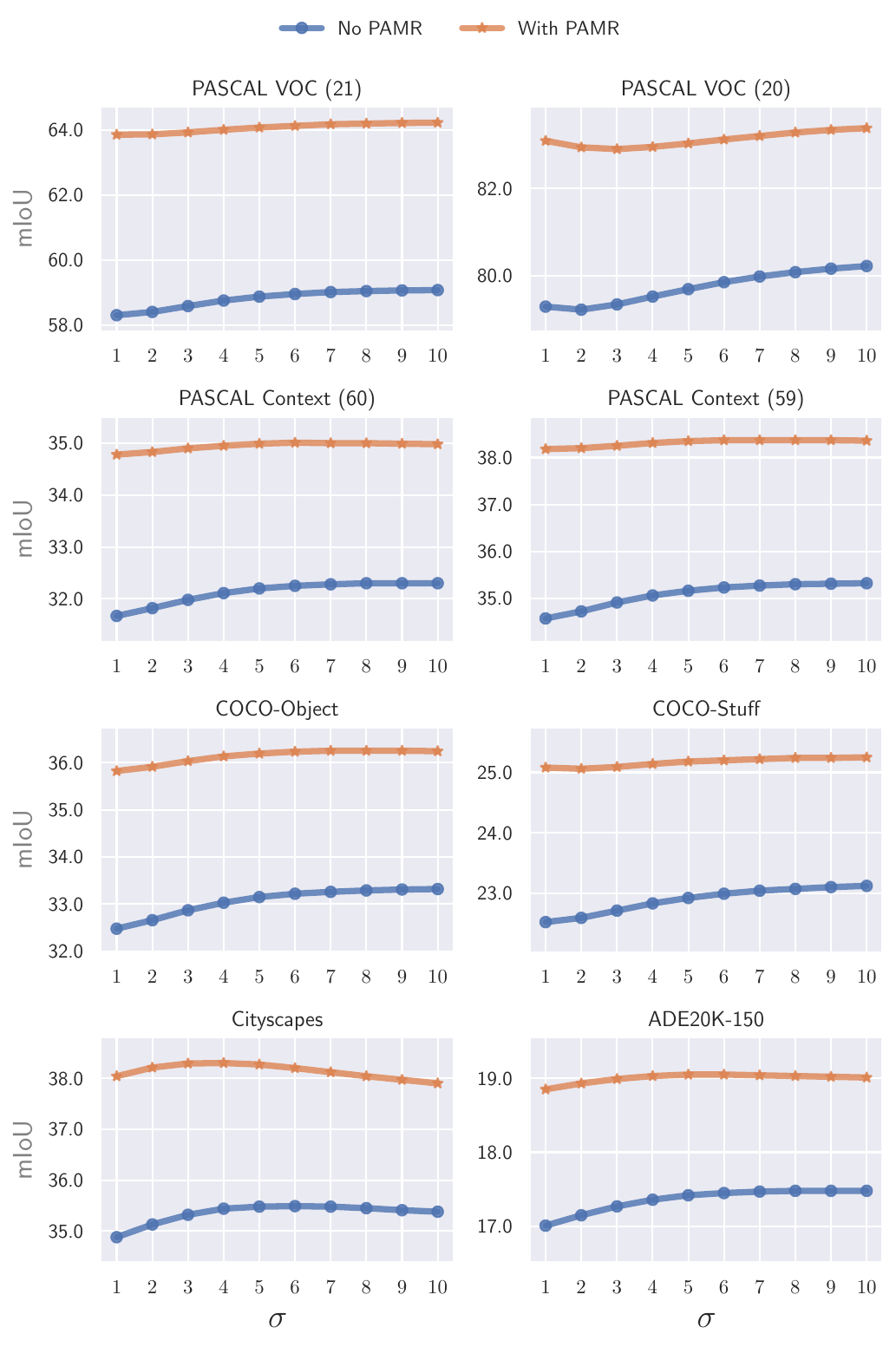}
    \caption{\textbf{Ablation study on the impact of $\sigma$.} We have provided results for both cases of using and not using post-processing, revealing consistent trends across both cases.}
    \label{app:fig:ablation-std-plot}
\end{figure*}

Although we employed a heuristic measure to determine $\sigma$, we provide in \cref{app:fig:ablation-std-plot} the impact of varying $\sigma$ values on test set performance. 
Please note that these experiments were conducted after deciding to use $\sigma = 5$, and whose goal is to demonstrate that 
$\textit{i)}$ our heuristic approach to find $\sigma$ provides indeed a good value; and 
$\textit{ii)}$ the performance across different datasets is not strongly sensitive to the hyperparameter $\sigma$.

\section{Visual examples} \label{app:visual}

\begin{figure*}
    \centering
    \includegraphics[width=0.85\textwidth]{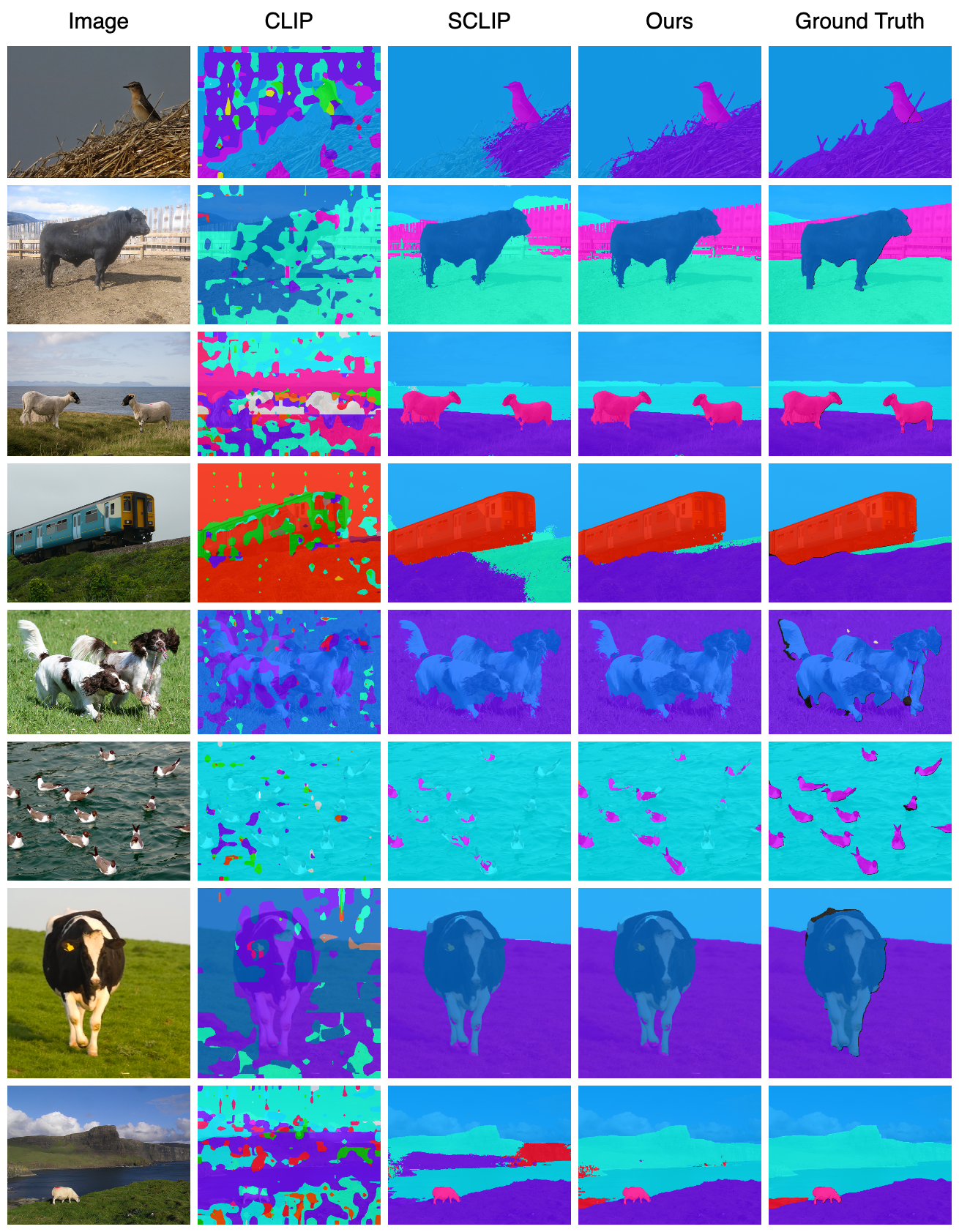}
    \caption{\textbf{Additional visual examples (segmentation maps) from PASCAL Context (59) \cite{pascalcontext}} for CLIP \cite{clip}, SCLIP \cite{sclip}, and our method.}
    \label{app:fig:pc59}
\end{figure*}

\begin{figure*}
    \centering
    \includegraphics[width=0.85\textwidth]{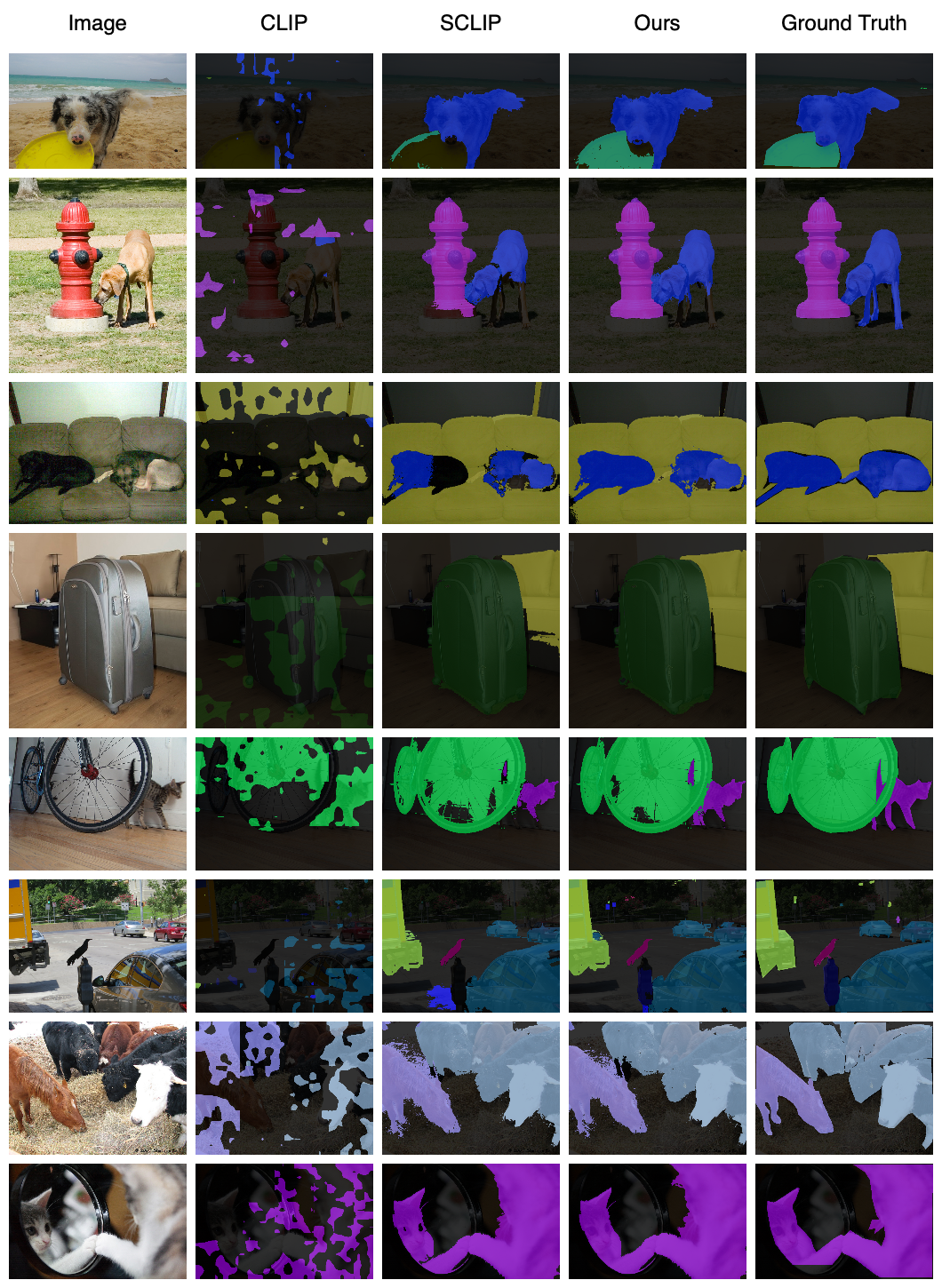}
    \caption{\textbf{Additional visual examples (segmentation maps) from COCO-Object \cite{actualcoco,coco}} for CLIP \cite{clip}, SCLIP \cite{sclip}, and our method.}
    \label{app:fig:cocoobj}
\end{figure*}

Additional visual examples can be found in \cref{app:fig:pc59} for PASCAL Context~(59)~\cite{pascalcontext}, and in \cref{app:fig:cocoobj} for COCO-Object~\cite{actualcoco,coco}. Upon reviewing the images in \cref{app:fig:pc59}, we can observe that SCLIP~\cite{sclip} often encounters difficulties in segmenting objects wholly and finding their boundaries (\eg, rows 1, 2, 4, and 8). We attribute this problem to SCLIP's failure to consistently incorporate information from surrounding patches.
Similar observations can be made for the first four rows of \cref{app:fig:cocoobj}. However, an interesting minute distinction between the methods emerges in the final row of the figure. Notably, the pixels representing the cat's eyes differ significantly from those of its skin, resulting in SCLIP failing to segment them as the same class. In contrast, \mbox{\ourmethod} attentively considers the surrounding context of the eyes, resulting in accurate segmentation.

\end{document}